\documentclass{article}

\PassOptionsToPackage{numbers, sort&compress}{natbib}

\usepackage[final]{neurips_2021}

\usepackage[utf8]{inputenc} 
\usepackage[T1]{fontenc}    
\usepackage{hyperref}       
\usepackage{url}            
\usepackage{booktabs}       
\usepackage{amsfonts}       
\usepackage{nicefrac}       
\usepackage{microtype}      
\usepackage{xcolor}         

\usepackage{mathrsfs}
\usepackage{mathtools} 

\usepackage{todonotes}

\usepackage{placeins}

\renewcommand{\tau}{\beta} 

\title{Two steps to risk sensitivity}

\author{%
  Chris Gagne \\
  MPI for Biological Cybernetics\\
   \\
  Tübingen, Germany\\
  \texttt{christopher.gagne@tuebingen.mpg.de} \\
  \And
  Peter Dayan \\
  MPI for Biological Cybernetics\\
  University of Tübingen\\
  Tübingen, Germany\\
  \texttt{dayan@tue.mpg.de } \\
}

\begin{document}

\maketitle

\begin{abstract}

Distributional reinforcement learning (RL) -- in which agents learn about all the possible long-term consequences of their actions, and not just the expected value -- is of great recent interest. One of the most important affordances of a distributional view is facilitating a modern, measured, approach to risk when outcomes are not completely certain. By contrast, psychological and neuroscientific investigations into decision making under risk have utilized a variety of more venerable theoretical models such as prospect theory that lack axiomatically desirable properties such as coherence. Here, we consider a particularly relevant risk measure for modeling human and animal planning, called conditional value-at-risk (CVaR), which quantifies worst-case outcomes (e.g., vehicle accidents or predation). We first adopt a conventional distributional approach to  CVaR in a sequential setting and reanalyze the choices of human decision-makers in the  well-known two-step task, revealing substantial risk aversion that had been lurking under  stickiness and perseveration. We then consider a further critical property of risk sensitivity, namely time consistency, showing alternatives to this form of CVaR that enjoy this desirable characteristic. We use simulations to examine settings in which the various forms  differ in ways that have implications for human and animal planning and behavior.

\end{abstract}

\section{Introduction}

Risk is  integral to decision making, arising whenever there are uncertain outcomes. It is especially critical when those outcomes are potentially calamitous, and plays an important role in psychiatric illness \citep{brand2006neuropsychological}. However,  psychological investigations into choice under risk (i) have yet to embrace the strong formal foundations being developed in finance, AI and machine learning; and  (ii) have mostly focused on one-shot decision tasks, despite the ubiquity of situations in the real-world that require planning, and growing interest in multi-step tasks for elucidating richer mechanisms of choice \citep{daw2011model,gillan2016characterizing}.

To address these lacun\ae, we consider a modern, theoretically well-founded coherent \citep{artzner1999coherent} risk measure called conditional value-at-risk (CVaR). This exactly quantifies the lower tail of possible outcomes -- those which are important for survival and also tend to drive our most persistent worries. CVaR has been applied to sequential decision problems, notably by means of distributional reinforcement learning (RL; \citep{bellemare2017distributional, dabney2018implicit, dabney2020distributional}). In this paradigm, the agent (or decision maker) estimates whole distributions for potential outcomes that can arise from its actions. Risk measures, such as CVaR, can be applied to these distributions to adjust decision making to any desired level of risk sensitivity.

In this paper, we combine CVaR with a distributional approach to examine risk sensitivity in the intensively-investigated two-step sequential decision task \citep{daw2011model}. We find that the choices of many individuals reflect a large degree of risk aversion. Moreover, we show that more standard analyses ascribe this to enhanced choice perseveration and/or reduced estimates for learning rate, thus misspecifying the effects of relatively  high levels of uncertainty. Incorporating risk sensitivity into choice, through CVaR or other risk measures, however, raises subtle issues regarding the consistency of choices over time \citep{boda2006time, artzner2007coherent, shapiro2009time, ruszczynski2010risk}. Such issues are well explored in psychology in the context of hyperbolic temporal discounting \citep{ainslie1992hyperbolic}, but their thorough investigation in finance and operations research has yet to permeate judgement and decision making research. We discuss how a direct incorporation of CVaR into distributional RL can lead to time inconsistency and point to two additional time-consistent approaches to CVaR for sequential decision making. Finally, we show how these three approaches can give rise to distinct patterns of behavior in a problem designed to tease them apart and which can serve as the basis for future empirical investigation.

\subsection*{Preliminaries: Conditional value-at-risk (CVaR)}

A risk measure $\rho(Z)$ maps a probability distribution of outcomes associated with a random variable $Z$ to a real number.  In distributional RL, $Z$ is typically a  discounted sum of rewards minus costs (i.e. the return). $\rho(Z)$ represents the risk inherent in the uncertainty about $Z$, and is often interpreted as the amount one is willing to pay to avoid adopting this risk.

\begin{figure}
  \centering
  \includegraphics[width=1.0\linewidth]{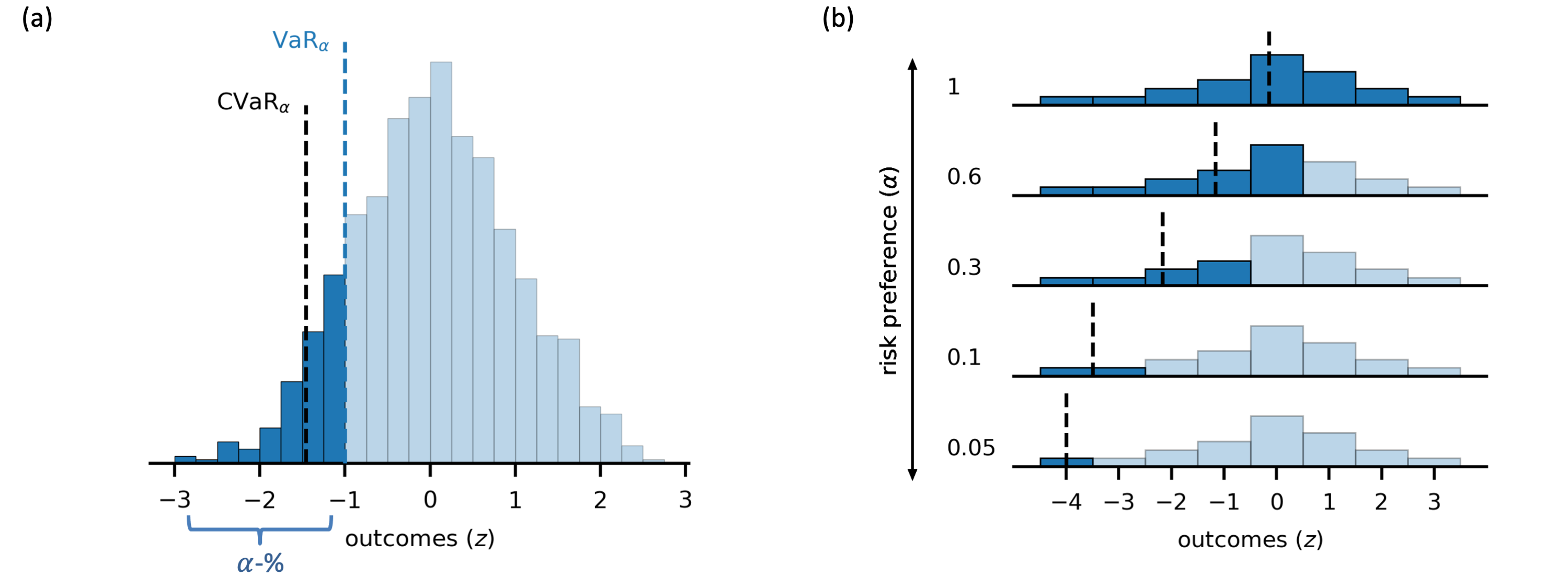}
  \caption{Conditional value-at-risk (CVaR). (a) $\text{CVaR}_\alpha$ is the average of the values in the lower $\alpha$-\% of a distribution, i.e. below the $\text{VaR}_\alpha$. (b) Adjusting $\alpha$ emphasizes the lower tail of the distribution more or less and sets the level of risk aversion. At $\alpha=1.0$, $\text{CVaR}$ is equal to the mean. As $\alpha$ decreases to $0$, it approaches the minimum.}
  \label{fig_cvar}
\end{figure}

Well known risk measures include the variance and the value-at-risk (VaR$_{\alpha}$), which is defined as: 
\begin{equation}
\label{eq_def_VaR}
\text{VaR}_\alpha(Z) = F^{-1}(\alpha) := \text{inf} \{z : F(z) \geq \alpha \}
\end{equation}
for a cumulative distribution function $F(z)$ and $\alpha$-quantile; see Figure \ref{fig_cvar}a. While variance and VaR are venerable risk measures, neither satisfy the full set of axiomatically desirable  properties associated with \emph{coherent risk measures} \citep{artzner1999coherent}. For instance, the VaR is not sub-additive (so fails to reward diversification), and variance is neither sub-additive nor positively homogeneous (as it changes non-linearly with the units in which costs or rewards are measured).

Conditional value-at-risk (CVaR) was introduced as a coherent risk measure \citep{artzner1999coherent}, and is particularly popular in finance, robotics, operations research, and recently machine learning and RL \citep{bauerle2011markov,morimura2012parametric,tamar2015optimizing,chow2015risk,chow2017risk,dabney2018implicit, keramati2020being,urpi2021risk}. For the lower tail of a continuous distribution, it is defined as the average of the values lower than the VaR (Figure \ref{fig_cvar}a):\footnote{For discrete random variables, alternative definitions such as $\text{CVaR}_\alpha(Z) := \text{sup}_\nu \{\nu - \frac{1}{\alpha}E[(\nu - Z)^+]\}$ \citep{rockafellar2002conditional} are used.}
\begin{equation}
\label{eq_def_CVaR}
\text{CVaR}_\alpha[Z] := E[Z | Z<\text{VaR}_\alpha(Z)] 
\end{equation}
$\alpha$ determines risk aversion by emphasizing the lower tail of the distribution more or less (Figure \ref{fig_cvar}b). 

That CVaR concentrates on the lower tail makes it particularly attractive for capturing aspects of normal and pathological reasoning in animals, whose lives often hang on  rather thin threads, and humans, who can catastrophize about diverse, unlikely, possibilities. But it is then particularly important to consider CVaR in sequential decision making, since this represents the ecological norm and is etched into the neural structure of decision making. We therefore start by examining CVaR in the simplest such problem that is widely studied in humans: the two-step task.

\section{Modeling risk sensitivity in human planning: CVaR in the two-step task}

\begin{figure}
    \centering
    \includegraphics[width=1.0\linewidth]{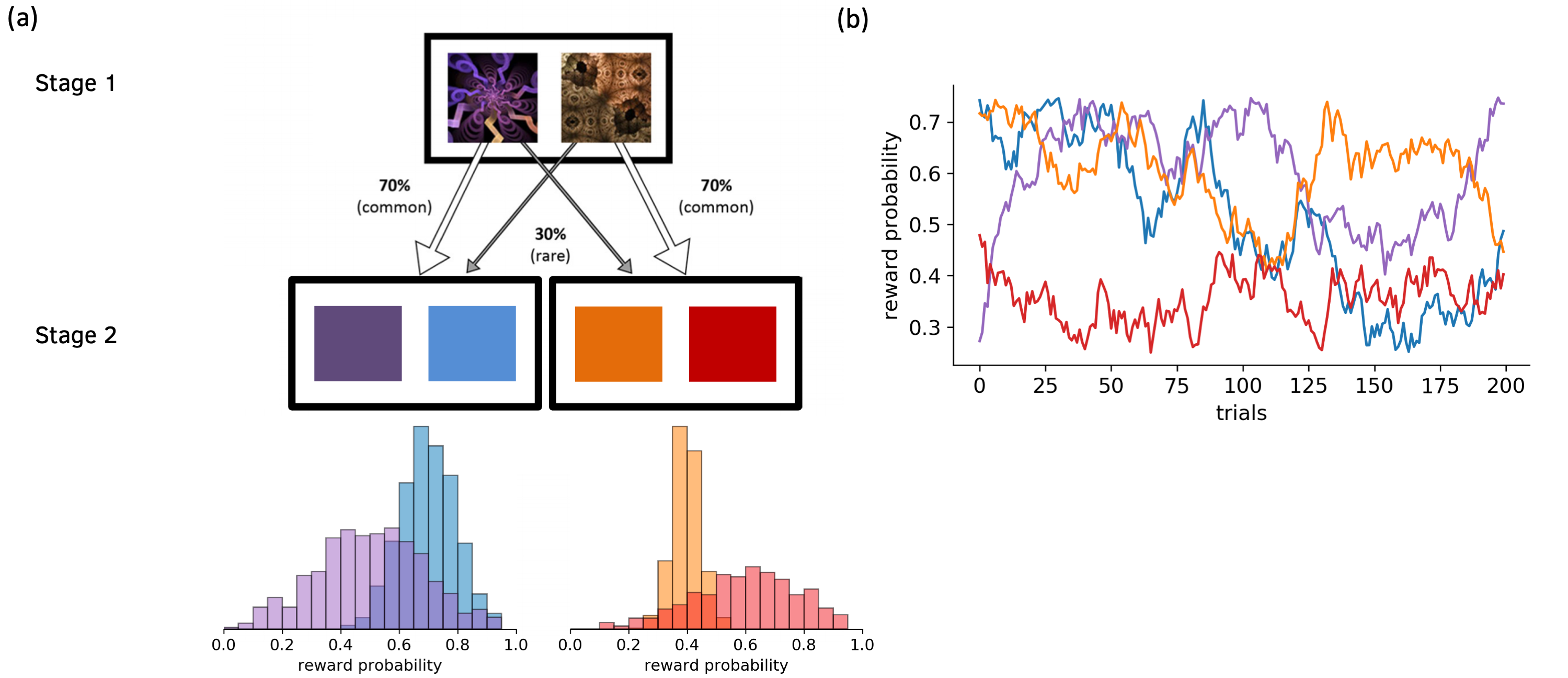}
  \caption{Two-step task. (a) On each trial, participants choose between two options (depicted as fractals) in stage 1 and then transition stochastically to one of two states (purple/blue or orange/red) in stage 2. They again choose between two options (the colored squares) and receive a binary reward. (b) This reward is sampled according to a probability that drifts randomly across trials (separately for each option). Distributions over reward probabilities are assumed by the model to be estimated for each option (depicted at the bottom in panel a).}
  \label{fig_twostep_task}
\end{figure}

The two-step task (Figure \ref{fig_twostep_task}) is a popular option for studying sequential choice  in humans (and animals) \cite{daw2011model}. It was originally designed to investigate model-based (MB) and model-free (MF) learning and planning, distinguishing the two by examining the consequences of progressive changes in the probabilities of rewards. However, these changes necessarily induce uncertainty -- which could affect risk-sensitive subjects. We therefore used a CVaR-based form of MB and MF reasoning to fit a very substantial dataset of human behavior in this task (out of more than 2000 participants in \citep{gillan2016characterizing}, the 792 who responded on every trial).

\paragraph{Task:}  On each of 200 trials, participants make  decisions at two successive stages or steps.  At the first stage, participants choose between two actions (the fractals in Figure \ref{fig_twostep_task}a). As shown by the arrows in the figure, depending on their choice, they then transition to one of two possible second stage states either 70\% or 30\% of the time. Each second stage state  has its own pair of options (the colored squares) between which the  participants must then choose. According to this second choice,  the participant receives a binary outcome, which is drawn according to a probability that drifts randomly across the trials (shown by color in Figure \ref{fig_twostep_task}b). Before the actual experiment, participants were given 40 trials of practice, during which they learned about the general structure of the task including the 70\%/30\% transition probabilities. This structure was not presented to participants explicitly, but participants were tested about their knowledge of the task and excluded if they failed this test. Participants were instructed about the drifting reward probabilities.

\paragraph{CVaR-based model:} We included CVaR in a model that is conventionally applied to the two-step task. We assume that participants learn a distribution of values \citep{bellemare2017distributional} associated with each of the four second-stage options using an approximate Kalman filter:
\begin{align}
 \mu_{t+1} = \mu_{t} + \lambda (r_t - \mu_{t}) \hspace*{.5in}
 \Psi^2_{t+1} = (1-\phi^2)\Psi^2_{t} + \eta^2 - \lambda \Psi^2_{t}
  \label{eq_var_update}
\end{align}
updating the mean $\mu_t$ of each distribution  on each trial using the observed outcome $r_t$ via a delta-rule with a participant-specific learning rate $\lambda$, 
and also updating the variance $\Psi_t$. When the outcome is not observed (i.e. when the participant chose a different option), the mean is updated towards 0.5 and the variance is updated without the last term in equation (\ref{eq_var_update}). Thus, the dispersion parameter $\eta^2$ controls the increase of the variance whether or not outcomes are observed, and the learning rate $\lambda$ controls the amount it decreases when they are. The term $(1-\phi^2)$ controls the asymptotic variance for unobserved outcomes (in relation to $\eta^2$) and was set based on the other two parameters (see later).

The CVaR$_{\alpha,t}(a)$ at risk preference $\alpha$ (estimated per participant) was calculated for each option $a$ using the mean $\mu_t(a)$ and variance $\Psi_t(a)$ on each trial $t$. Note that because this variance represents uncertainty about the reward probabilities themselves, CVaR here captures both ambiguity- as well as risk-sensitivity; however, to keep terminology consistent with the finance and machine learning communities \cite{fan2015dynamic,sharma2019robust,ahmadi2020risk,rigter2021risk}, we simply refer to this as to risk-sensitivity.

Equipped with these CVaR$_{\alpha,t}(a)$ values, participants' second stage choices were modeled using a soft-max choice rule:
\vspace{-2pt}
\begin{equation}
 \label{eq_choice_rule1}
 P(2^{nd} \text{ stage choice}\!=\!a) \propto \exp(\tau^{2nd} \text{CVaR}_{\alpha,t}(a))
\end{equation}
where the parameter $\tau^{2nd}$ controls the relative stochasticity/determinism.

Decisions between the two first-stage options were modeled as involving a combination of  model-free (MF) and model-based (MB) approaches to value estimation -- both of which were modified to include CVaR.  MF estimates were calculated using the same formul\ae, learning rate $\lambda$ and dispersion parameter $\eta^2$ as at the second stage to learn an additional pair of means and variances based on the actual outcomes received in the second stage. For model-based estimation, the 70\%/30\% transition probabilities were used to calculate a mixture distribution for each of the top stage actions, from which the CVaR was calculated directly. CVaR$_{\alpha,t}(a)$ for the model-free and model-based first-stage distributions again determined the choice probability through a soft-max choice rule:
\begin{equation}
 \label{eq_choice_rule}
 P(1^{st} \text{stage choice}\!=\!a) \propto \exp(\tau^{MB} \text{CVaR}^{MB}_{\alpha,t}(a)  + \tau^{MF} \text{CVaR}^{MF}_{\alpha,t}(a) + \tau^{\text{sticky}}\delta_{a,a_{t-1}})
\end{equation}
Parameter $\tau^{\text{sticky}}$ was included to capture the tendency of participants to repeat (or to switch) the previously chosen action regardless of its value.  

\paragraph{Parameter estimation:} The $7$ parameters of the CVaR-model were estimated for each participant: CVaR-based risk-sensitivity  $\alpha \in [0.1,1]$,  learning rate $\lambda \in [0.01,0.99]$, dispersion  $\eta^2 \in [0.001,0.09]$, perseveration  $\tau^{\text{sticky}} \in [0,20] $ and three inverse temperature parameters $\tau^{2nd}, \tau^{MB}, \tau^{MF} \in [0,30]$. Parameters were estimated in Python using L-BFGS-B. Parameter recovery analyses were conducted to investigate parameter estimability. Preliminary recovery simulations suggested that estimating both $\eta^2$ and $(1-\phi^2)$ was difficult, so we determined the value of $(1-\phi^2)$ that would pin the asymptotic variance to 0.1. This was done so that a never-chosen option with an estimated mean 0.5 would have a CVaR of 0 at the lower-bound value for $\alpha$ (i.e. $0.1$), which is appropriate since the outcomes themselves were between 0 and 1. The learning rate lay between 0.01 to 0.99 and was additionally constrained such that $\lambda < (1-\phi^2)$. With these constraints, we ran further parameter recovery analyses, generating new data based on participants' parameter estimates and re-estimating the model; a rank-correlation of 0.71 between the generative and recovered CVaR $\alpha$ parameter indicated moderately good estimability/identifiability; Supplemental Figure 1. 

\paragraph{Baseline and alternate models:}

Setting $\alpha=1$ arranges for CVaR choices that depend only on expected values (in which case the variance estimation can also be removed, because it does not influence choice probability). This mean-model is used as a baseline against the risk-sensitive CVaR$_{\alpha<1}$ model; it is very similar to typical models used in the two-step task \citep{gillan2016characterizing, daw2011model}, but is more directly comparable to its risk-sensitive counterpart. We provide a more detailed description of the differences between the current and previously used model in Appendix A.2. We also tried two forgetful beta-binomial models (one with and without risk-sensitivity) since rewards are binary, and a risk-seeking version of the CVaR-model, which emphasized the upper rather than lower tail. However, these models fit participants' data less well, so we omit them for the sake of parsimony.

\section*{Results}

\begin{figure}
  \centering
  \includegraphics[width=1.0\linewidth]{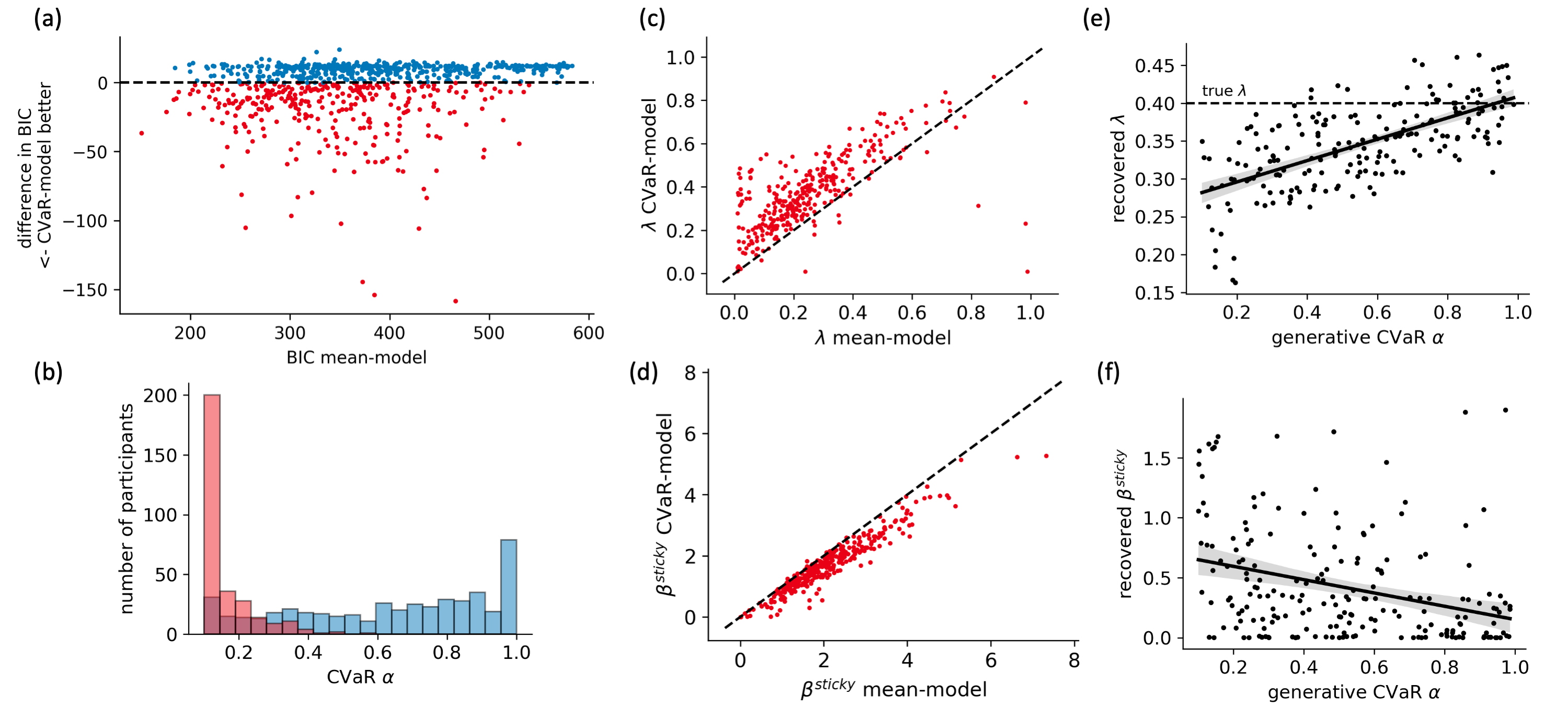}
  \caption{Two-step task modeling results. (a) Many participants are better fit by the CVaR-model than the mean-model, and (b) are substantially risk averse, as estimated by the CVaR $\alpha$ parameter. (c-d) Including risk sensitivity increases/decreases estimates of learning rate and perseveration, respectively, in these participants. (e-f) Simulations show that fitting a risk-agnostic model to risk-sensitive behavior can lead to underestimated learning rates (e) and/or apparent perseveration (f). Shading around the regression line in panels e-f indicate 95\%-confidence intervals. Note that 3 outliers located near (x=12, y=10) were removed from panel d.}
  \label{fig_twostep_results}
\end{figure}

We first compared the the mean-model (CVaR$_{\alpha=1}$) to the risk-sensitive model (CVaR$_{\alpha<1}$). Using BIC to account for the two extra parameters,  the risk-sensitive model was slightly preferred overall (avg. BIC=368.8 versus BIC=371.5; diff=-2.67, t(790)=-3.54, p=0.004), albeit in a minority (40.2\%) of participants. For many such subjects, however, the improved fit was substantial (red points in Figure  \ref{fig_twostep_results}a); and many of them had $\alpha<0.2$ (Figure  \ref{fig_twostep_results}b), indicating substantial aversion to risk.

To investigate the improved fit and for insight into the choice characteristics associated with this risk aversion, we analyzed how other parameters altered from the $\alpha\!=\!1$ case for significantly risk-averse subjects. There were two related changes: participants' estimated learning rates were higher in the CVaR-model (Figure \ref{fig_twostep_results}c), and their estimated perseveration parameters were  lower  (Figure \ref{fig_twostep_results}d). One potential reason for these is that participants tended \emph{not} to prefer less frequently chosen (and thus more uncertain) options, even when the more certain option is apparently worse.

\begin{figure}
  \centering
  \includegraphics[width=1.0\linewidth]{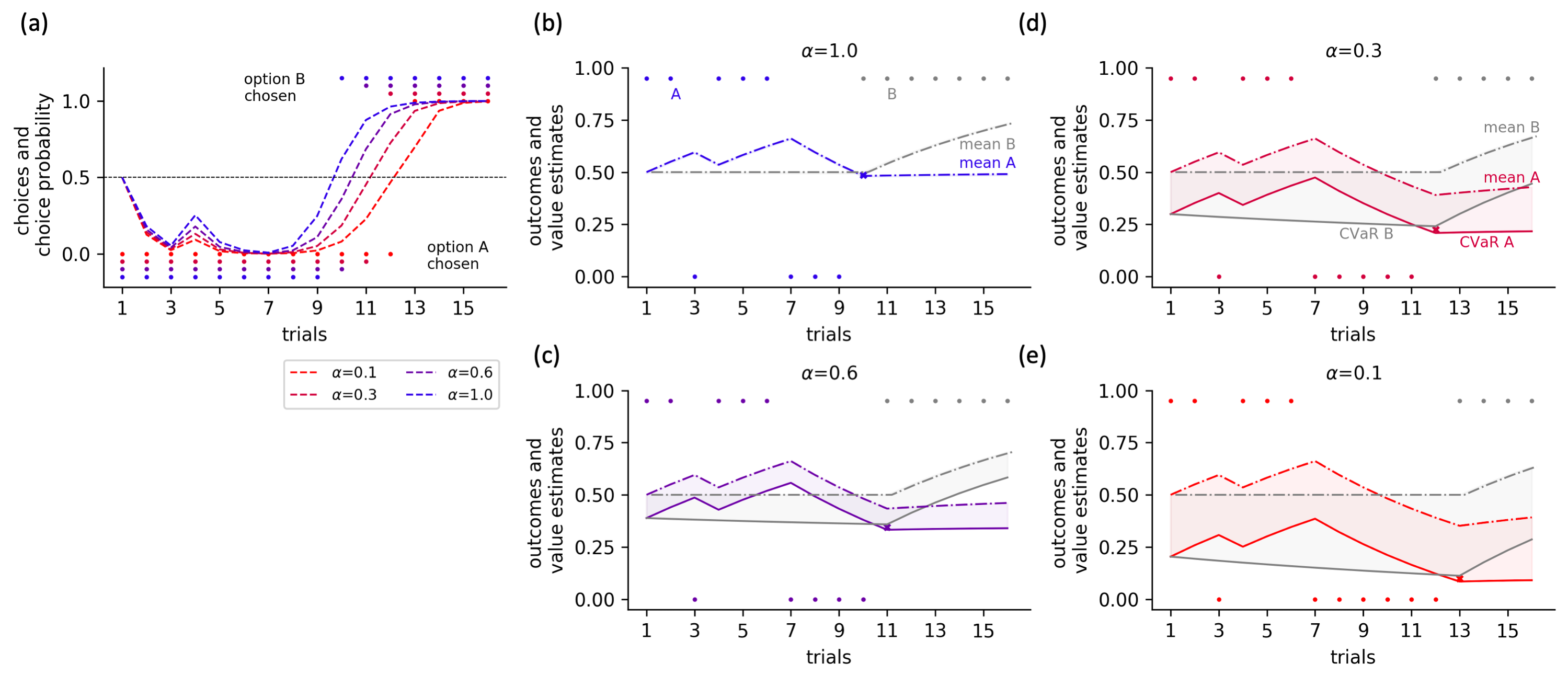}
  \caption{CVaR-based perseveration. (a) The CVaR-model is simulated at four  values of $\alpha$ on a series of choices between options A and B. For low $\alpha$ (high risk sensitivity), the model chooses option A for more trials, despite receiving $0$ outcomes after trial 7; binary outcomes are shown as dots in panels b-e. (b-e) The switch from A to B occurs when the model's estimate for the CVaR (or mean for $\alpha=1.0$) of  A (in color) goes below that of  B (in gray). The crossing point occurs later for lower levels of $\alpha$ due to uncertainty around the mean which is greater for  B.}
  \label{fig_twostep_simulation}
\end{figure}

To demonstrate this directly, we simulated choices from the CVaR-model at increasing levels of risk sensitivity ($\alpha \in [1.0, 0.6, 0.3, 0.1]$) in response to a predetermined set of outcomes, with the other parameters held constant (Figure \ref{fig_twostep_simulation}). Option A is chosen by design for the first 6 trials, but the model is then allowed to switch to an option B after observing what it thinks is a sufficient number of negative outcomes. For $\alpha\!=\!1$, this switch occurs when the mean estimate for option A crosses the mean estimate for option B at 0.5 (trial 10 in panel b). For $\alpha\!<\!1$ (panels c-e), this switch occurs when the CVaR$_{\alpha<1.0}$ estimate for the options cross. Due to the uncertainty around the mean and the fact that the model is more uncertain about option B than A, the crossing point occurs later for lower $\alpha$ (i.e. on trials 11, 12, and 13). Thus, what looks like a tendency to stick with an apparently worse option, here arises from an aversion to uncertainty. For $\alpha\!=\!1$, this can only occur if the learning rate is low (so the value of the chosen option does not decrease too much upon experiencing an unfortunate outcome), and/or with a high perseveration parameter. But, as a consequence, genuine risk aversion (or uncertainty aversion) might be misattributed to one of these parameters -- this is shown directly in Figure \ref{fig_twostep_results}e;f where
generating choices from a non-perseverative CVaR-model with
low $\alpha$ leads to inferred learning rates that are too low
(panel e) and inferred non-zero perseveration
(panel f). Note, though, that in participants' data, some perseveration remains even with CVaR$_{\alpha<1}$, as removing perseveration from the CVaR-model (i.e. setting $\tau^{\text{sticky}}=0$) led to worse BIC values.

\section{Three approaches to sequential risk and time (in)consistency}

One of the most important issues for sequential decision making is time consistency -- that the choices the decision maker at time $t$ assumes will be executed at time $t+1$ are indeed carried out.\footnote{An intimately related, yet differently defined, notion of time consistency involves the consistency of successive risk evaluations of a stochastic process rather than choices per se. Of course, inconsistent evaluation can lead to inconsistent choices. See \citep{de2016building,shapiro2014lectures,rudloff2014time} for a more in-depth discussion.} Inconsistency, famously caused by hyperbolic discounting \citep{ainslie1992hyperbolic}, can lead to reneging on past resolutions or require potentially sub-optimal and expensive commitment behavior to prevent that from occurring. Another, less well-known, route to time inconsistency comes from mishandling risk sensitivity \citep{kreps1979,epstein2003recursive,boda2006time,artzner2007coherent,roorda2007time,shapiro2009time,ruszczynski2010risk,rudloff2014time,pflug2016time,shapiro2014lectures,pichler2018risk,schur2019time,majumdar2020should} -- and, indeed, different variants of CVaR have been developed to deal with this issue. The two-step task lacks sufficient temporal complexity to make this issue clear. Therefore in this section we examine it in greater depth, and, using simulations, suggest a task that highlights the differences between these CVaR variants.

\FloatBarrier

\begin{figure}
  \centering
  \includegraphics[width=1.0\linewidth]{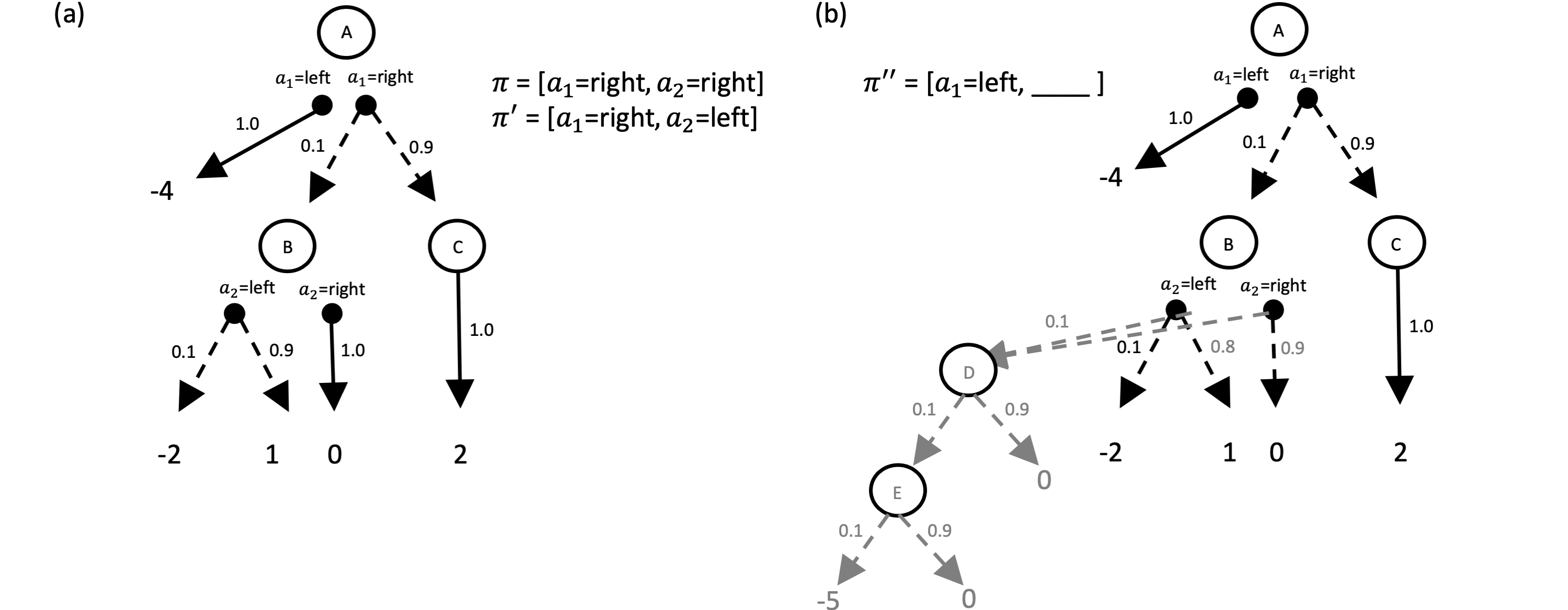}
  \caption{Three approaches to CVaR in a two-stage problem. (a) A choice of left or right can be made in both state A and B (if state B is visited). Stochastic transitions are denoted by dashed lines (with the probabilities shown), and the possible terminal outcomes are [-4, -2, 1, 0, 2]. The fixed approach to CVaR (discussed in the text) would choose policy $\pi^\prime$ at state A, but policy $\pi$ at state B, illustrating issues with time consistency. The precommitted approach would consistently choose policy $\pi^\prime$. (b) In an adjusted problem, additional state transitions (colored in gray) are appended to state B. The nested approach to CVaR (with $\alpha=0.1$) chooses policy $\pi^{\prime\prime}$, taking a sure loss of $-4$, despite the only extremely remote chance of getting $-5$.}
  \label{fig_CVaR_tree}
\end{figure}

\subsection*{Fixed, precommitted, and nested CVaR}

In the two-step task, we followed recent work on distributional RL  \citep{dabney2018implicit} in assuming that  participants applied CVaR at the same risk preference ($\alpha$) to the estimated value distributions at both the first and second stage. We call this the fixed approach or fCVaR. 
To see why this approach is time-\emph{in}consistent, consider calculating the optimal fCVaR$_{\alpha=0.1}$ policy for the simplified two-stage choice scenario depicted in Figure \ref{fig_CVaR_tree}a. 

From the perspective of the top state A, consider the two policies $\pi\!=\!\{a_1\!=\!\textit{right},a_2\!=\!\textit{right}\}$, which leads to a distribution of outcomes $p\!=\!\{0.1\!:\!0,0.9\!:\!2\}$ and $\pi'\!=\!\{a_1\!=\!\textit{right},a_2\!=\!\textit{left}\}$, which leads to a distribution of outcomes $p'\!=\!\{0.01\!:\!-2,0.09\!:\!1,0.9\!:\!2\}$. Just considering these overall distributions, $\pi$ has a $\text{CVaR}_{0.1}$ of $0$ and $\pi'$ has an $\text{CVaR}_{0.1}$ of $0.7$, so $\pi'$ would be preferred. However, if the agent takes action $a_1\!=\!\textit{right}$ and arrives at state B, the remainder of $\pi'$ (i.e. action $a_2\!=\!\textit{left}$) now looks worse than the remainder of $\pi$ (i.e. $a_2\!=\!\textit{right}$), with $\text{CVaR}_{0.1}$'s of $-2$ and $0$, respectively; as a result, the fCVaR agent will defect on its original plan $\pi'$.

One way to circumvent this issue with time consistency is to skirt dynamic evaluation altogether and precommit to evaluating risk with respect to only one stage or time-point (the start being most natural) -- thus enforcing a commitment $\pi'$ at state A. One way to make this contract is to change the value of $\alpha$ after each transition in the light of the probability of its happening. For instance, the decision in $\pi'$ made at state A to go \textit{left} rather than \textit{right} at state B was based on considering all  the outcomes in state B (i.e. the decision was based considering $\text{CVaR}$ at $\alpha\!=\!1.0$ rather than $\!=\! 0.1$). This dynamic adjustment of $\alpha$ (in this case from $0.1$ in state A to $1.0$ in state B) prevents time inconsistency by allowing later risk evaluations and decisions all to be coordinated with respect to the single risk preference at the start stage \citep{pflug2016time}.

A second way to deal with time consistency is to evaluate risk dynamically using a series of nested one-step (conditional) risk measures \citep{ruszczynski2006conditional,ruszczynski2010risk}:
\begin{equation}
 \label{eq_nested_risk_measures}
\rho_{k,N}(R_k,\dots,R_N) = R_k + \rho_k(R_{k+1} + \rho_{k+1}(R_{k+2} + \cdots + \rho_{N-2}(R_{N-1} + \rho_{N-1}(R_N))\cdots)),
\end{equation} 
where $R_k$ is the reward (or cost) for time step $k$.

If CVaR$_{\alpha}$ with the same value of $\alpha$ is used for each conditional risk measure $\rho_k$, this is known as nested or nCVaR$_{\alpha}$ \citep{pflug2007modeling}. Although like fCVaR, it keeps risk preferences fixed across time, it applies the CVaR at each stage to subsequent CVaR evaluations (which themselves are random \emph{scalars}) rather than to the \emph{full distribution} of future random costs or rewards under the policy lower in the tree.

One consequence of this nesting or compounding of risk evaluations, is that nCVaR can sometimes become much more conservative than the other two approaches. To see this, consider panel b in Figure \ref{fig_CVaR_tree}. Here, the same decision tree is appended with a set of alternative transitions (depicted in gray) from state B. Now, each possible action is additionally associated with a 10\% chance of ending up in state D, which either returns 0 with probability $0.9$ or makes a transition to yet another state E with probability $0.1$. The outcomes in this state are either $-5$ (the worst possible outcome) or again $0$. Importantly, from the perspective of state B, the $-5$ is so remote (having a probability of merely $0.001$) that it barely impacts the distribution there (and therefore pCVaR and fCVaR). However, due to the nested structure of nCVaR (at $\alpha=0.1$), the $-5$ is propagated backwards first to E then to D, and then in fact, all the way back to the top state A. As a consequence, the nCVaR agent would then rather choose $\pi^{\prime \prime} = \{a_1=\textit{left}, \text{N/A}\}$, for a certain loss of -4.

Of course at $\alpha=1.0$, all three approaches are equivalent to using the expected value, which is time consistent. As $\alpha$ approaches $0$, all three approaches again become equivalent, but to the worst-case risk measure \citep{coraluppi1999risk}, which is also time consistent.

\subsection*{The three approaches in a gridworld}

\begin{figure}
  \centering
  \includegraphics[width=1.0\linewidth]{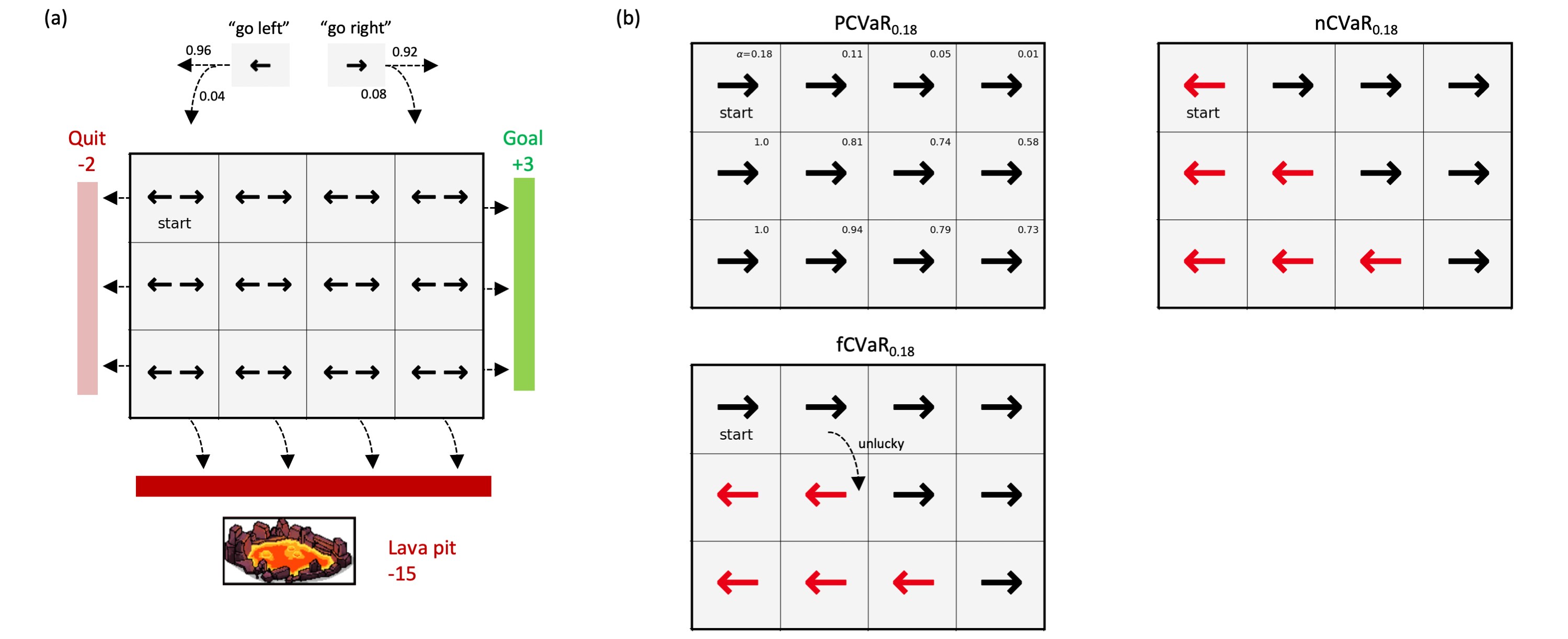}
  \caption{Three approaches to CVaR in a gridworld. (a) The agent starts in the upper left corner and can choose to go left or right in each state. Actions result in stochastic transitions (as depicted at the top). Exiting the map on the left, right, or bottom results in an outcome of -2, +3, -15. (b) The optimal policy is shown for each of three approaches, highlighting their distinct behaviors.}
  \label{fig_CVaR_gridworld}
\end{figure}

As noted, the two-step task is too simple to surface these issues (evaluating the other forms of CVaR leads to equivalent results on the 792 subjects). Thus, we used the simulations in Figure \ref{fig_CVaR_gridworld} to highlight their differences as the basis of possible future tests.

Consider an agent (or decision maker) who starts in a gridworld in the top left corner and can move either right or left. Exiting the gridworld on the right hand side informally represents a goal and is associated with a +3 reward, while exiting the map on the left hand side represents quitting and a loss of -2. The agent's actions are stochastic with the possibility of moving downwards with some error probability. If the agent falls off the bottom of the gridworld, a substantial loss of -15 is incurred (schematized by the lavapit). The right action has twice the error probability of the left action, meaning that  heading towards the goal is riskier than attempting to quit; this leads to interesting differences for the three approaches.

Optimal polices for the pCVaR, fCVaR and nCVaR were calculated using dynamic programming equations (see Appendix B; based on \citep{chow2015risk,ruszczynski2010risk}) for various levels of $\alpha$ (for pCVaR, which requires interpolating between different values of alpha, we used 21 log spaced points, following \citep{chow2015risk}). The optimal policies for each approach to CVaR at a moderate-to-high level of risk (i.e. $\alpha=0.18$) are depicted in Figure \ref{fig_CVaR_gridworld}b. 

For pCVaR (with a start state risk of $\alpha=0.18$), the optimal policy in every state is to head towards the goal. This is the case even in the bottom row, where rightward actions are much more risky due to the proximity of the lava pit. In fact, if the agent started in the bottom row with the same $\alpha=0.18$, it would head left in an attempt to quit. However, since the pCVaR approach coordinates risk with respect to the start state and these bottom states are part of the start state's lower tail, the pCVaR agent knows that it is better to make ‘riskier’ decisions here for the sake of its former self.\footnote{This increase in risk sensitivity can be seen in the average values of the adjusted alphas in each state; shown in the upper right hand corner of each grid cell.} Indeed, doing so yields a higher start state CVaR than abandoning its plans. In contrast, abandoning the pursuit of the goal is exactly what the fCVaR agent does, as it re-evaluates risk using $\alpha=0.18$ at every state; i.e. it heads towards the left in rows 2 and 3 after getting knocked off course. Similarly to fCVaR, the nCVaR agent also re-evaluates risk using $\alpha=0.18$ at every state. However, it chooses to quit from the start due to the nesting of risk (risk evaluated on top of future risk evaluations) -- even from the start, the chances of the distal lava pit looms larger than it does for the other two approaches (at the same nominal level of alpha). At lower and higher values of alpha, the behavior of the three approaches more closely aligns, either quitting from the start or pursuing the goal from every state.

\section{Discussion and related work}

We first showed that adopting the modern risk measure CVaR \citep{artzner1999coherent} in a form based on ideas from distributional RL \citep{dabney2018implicit} suggests that a large minority of healthy volunteer subjects exhibit quite significantly risk-averse behavior. This was true even in a simple two-step task for extremely low stakes and with only rather limited amounts of uncertainty. That this risk aversion masqueraded as enhanced perseveration and slowed learning is a reminder of the complexities of building models of behavior, and an invitation to consider risk in models of other common tasks. These effects of risk in behavior can be complemented by considerations of its effects in the sort of off-line planning that has recently been suggested to model rumination in anxiety disorders, as subjects struggle to find ways of mitigating potential future threats \citep{gagne}. That distributional RL has recently been suggested as a way of understanding facets of the activity of major neural systems involved in processing affective outcomes \citep{dabney2020distributional} offers a highly attractive link to understanding risk aversion in animals (and also tempts us to consider the role of other neuromodulators that have been implicated in representing and processing uncertainty; \citep{angela2005uncertainty}).

Since pathological effects of uncertainty play a malign role in the effects of many psychiatric conditions \citep{brand2006neuropsychological}, it would be of great interest to design tasks, for instance based on the gridworld of section~3, that decompose various of its different facets to help determine which might be responsible. Indeed, this task can be seen as a structurally rich form of a popular method for assessing impulsivity -- the balloon analogue risk task (BART; \citep{lejuez2002evaluation}). Many potential variants would be of interest -- for instance studying forms of precommitment by allowing subjects to buy a form of ‘insurance’ at the outset against one or more unlucky downwards transitions. We could modify the task further with intermediate rewards and punishments to examine the observation that individuals only take on greater risk after a loss, if it is not yet realized (in our terms, before reaching one of the two sides of the grid) \citep{imas2016real}. Then, we might expect precommitments to be abandoned and risk to be re-evaluated to the degree to which previous outcomes seem irreversible or the current situation sufficiently distinct from the one in which the committment was made.
One could examine consistency between subjects' choices in the environment and their willingness to pay to protect themselves ahead of time. Furthermore, richer distributions of possible outcomes and a diverse set of navigation environments, with features that differentiate the three approaches, could be used to reverse engineer uncertainty calculations in the brain given stable risk preferences. 

One particular facet of uncertainty that we did not explore is that, in some circumstances, it is possible to collect more information that reduces it to acceptable levels.  Indeed, modern risk measures have recently been applied applied in settings with model uncertainty (i.e. POMDPs and Bayes adaptive MDPs, \cite{fan2015dynamic,sharma2019robust,ahmadi2020risk,rigter2021risk}). Information gathering, as explored in the sequential information sampling task \citep{clark2006reflection} has been of particular interest in obsessive compulsive disorder, where subjects have been algorithmically modelled as exhibiting less urgency to make up their minds \citep{hauser2017increased}; it would be interesting to model them computationally as being more risk averse. Similarly, if trust is seen as willingness to risk vulnerability to others \citep{mayer1995integrative,rousseau1998not}, then risk sensitivity could be an important factor in social pathologies such as borderline personality disorder. 

We showed that fCVaR, although intuitive, is not time consistent. The other forms of precommitted pCVaR and nested nCVaR are, which provides them with more attractive formal properties. In fact, one can see fCVaR as living inbetween pCVaR and nCVaR (rather literally, in the problem shown in Figure~\ref{fig_CVaR_gridworld}b). There are other potential interpolants -- for instance, cases in which the updating of $\alpha$ following lucky or unlucky transitions (which justifies a form of gambler's fallacy; \citep{gagne}) is incomplete (or perhaps asymmetric). Such incompetent calculations have been suggested as underlying psychiatric conditions themselves \citep{huys2015decision}.

Finally, although we focused on CVaR, there are many other risk measures that satisfy the coherence axioms, and indeed other more stringent conditions. Adding the requirements of comontonicity and law invariance lead to the class of \emph{distortion risk measures} \citep{wang1996premium,wang2000class,follmer2016stochastic} (or equivalently, \emph{spectral risk measures} \citep{acerbi2002spectral}), which apply a distortion function to cumulative probabilities. This allows them to be linked directly to the \emph{dual theory} of choice \citep{yaari1987dual}, therefore inheriting an additional set of rational choice axioms (i.e. the axioms of expected utility theory, with an alternative independence axiom)\citep{wachter2013consistent}. CVaR itself is part of this class, and can be used as a basis to construct all other members \citep{follmer2016stochastic,majumdar2020should}. Interestingly, the probability weighting function from cumulative prospect theory \citep{kahneman1979,tversky1992advances}, a popular model in psychology, can be considered a distortion risk measure, even though the full prospect theory adds reference dependence and loss aversion. While prospect theory has been well validated for single decisions, there is substantial opportunity for theoretical and empirical investigations of its realization in sequential decision-making.

\section{Broader Impact}

Coherent risk measures, despite their widespread use in formal applications and favorable theory, have yet to permeate fully psychological and neuroscientific models of decision making, especially for sequential problems. We take early steps in this direction using a widely-recognized and commonly used experimental paradigm, and a psychologically relevant coherent risk measure, CVaR. In doing so, we highlight a key issue that can arise, namely time inconsistency, with a naive application of risk measures to sequential choice, and discuss two alternative (time-consistent) approaches. We expect that bringing awareness to this issue will lead to interesting future discussions and empirical tests in psychological and neuroscientific communities. We do not anticipate our research to selectively impact some groups at the expense of others. Our study had the limitation of not examining how subjects might compute these various forms of risk sensitivity in a neurobiologically credible manner, or showing that the risk aversion that we inferred from behaviour in the two-step task would generalize to other choices that the subjects might make.

\section*{Acknowledgements}

The authors have no competing interests to disclose. We thank Fabian Renz for his contributions to a preliminary analyses of the two-step task and helpful discussions. CG and PD are funded by the Max Planck Society. PD is also funded by the Alexander von Humboldt Foundation. 

Code is available at \href{https://github.com/crgagne/twosteps_neurips2021}{https://github.com/crgagne/twosteps\_neurips2021}.

\bibliography{references}

\end{document}


\appendix

\section{Further details for the two-step task analyses}

\FloatBarrier

\subsection{Parameter recovery} 
\begin{figure}[h]
  \centering
  \includegraphics[width=1.0\linewidth]{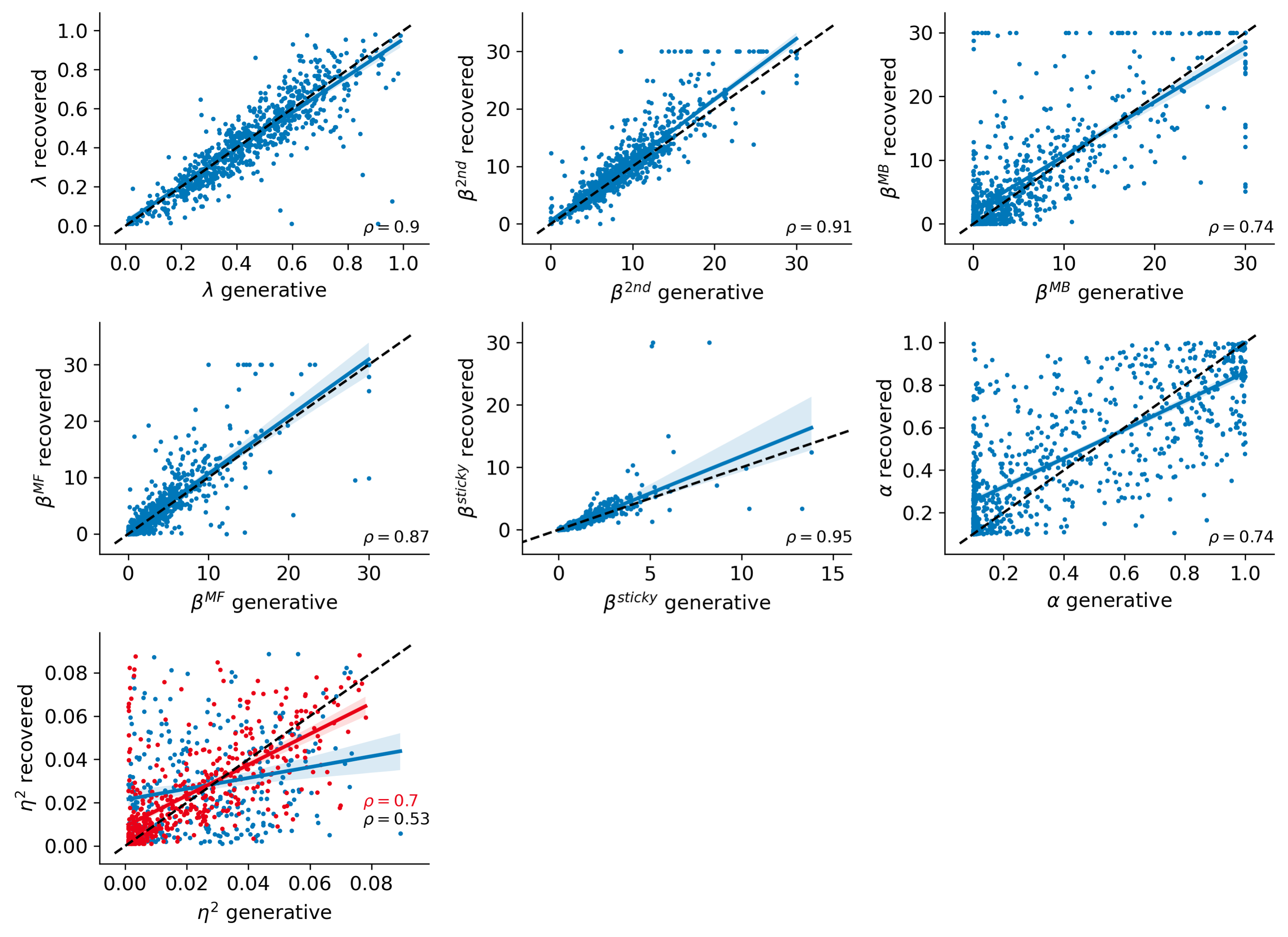}
  \caption{Parameter recovery analysis. (See supplemental text for details)}
  \label{fig_param_recovery}
\end{figure}

To assess the estimability/identifiability of the model parameters, we performed a parameter recovery analysis (Supplemental Figure~\ref{fig_param_recovery}).  New data were generated from the model based on participants' parameters (referred to as ‘generative’) and the model was re-estimated on these data. The rank correlation between the generative and re-estimated (or ‘recovered’) parameters was greater than $0.74$ (for all parameters except for the dispersion parameter, $\eta^2$), indicating moderately good estimability/identifiability. The rank correlation for $\eta^2$ was lower ($0.53$), but this is expected, because its identifiability depends on the level of $\alpha$; at $\alpha=1.0$, for instance, choices do not reflect any outcome uncertainty, making  $\eta^2$ unidentifiable. However, the estimability/identifiability of $\eta^2$ was similar to the other parameters (i.e. a rank correlation of $0.7$) in participants with $\alpha<0.5$ (shown in red). Correlations were all highly significant ($p<1e{-10}$), as shown by the $95\%$-confidence intervals around the regression lines.

\newpage
\new{
\subsection{Comparison of mean-model to previous model used in two-step task}

The mean-model that we used was modified from that in \citep{gillan2016characterizing} to allow for a more direct comparison to the risk-sensitive model. There are two small differences between our model and theirs. One difference is that the model in \citep{gillan2016characterizing} uses both rewards and the intermediate values from stage 2 to estimate the values in stage 1 (equation 5). We found this to be unnecessary, as it increased (worsened) the BIC in almost all (>95\%) participants. The second difference is one of re-parameterization. The rule that we use to update the mean values (equation 3) keeps the values between $[0,1]$, while the update rule used in \citep{gillan2016characterizing} allows them to go above $1$. We made this adjustment so that the value estimate had the semantics of a probability estimate. We found that this re-parameterization made very little difference in model fit (average BIC of 371.5 for ours and 373.9 for theirs).
}

\subsection{Details for the simulation shown in Figure 4} As described in the main text, choices were simulated from the CVaR-model at increasing levels of risk sensitivity ($\alpha \in [1.0, 0.6, 0.3, 0.1]$), in response to a predetermined set of outcomes. The model was forced to choose option A for the first six trials and the outcomes received were $[1,1,0,1,1,1]$; from trial seven onward, A always resulted in outcome $0$ and B in outcome $1$. The other parameters were held constant, with learning rate $\lambda=0.1$, $\eta^2=0.003$, and $\tau^{2nd}=30$. The initial mean for the value estimates was set to $50\%$ and its initial variance to $0.03$. Since the model was only simulated in response to two options at a single (second-stage) state, parameters that were only relevant to the first-stage decisions were not used (i.e. $\tau^{\text{sticky}}=0$, $\tau^{MB}=0$, $\tau^{MF}=0$).

\newpage
\section{Three Approaches to CVaR: Dynamic Programming Equations}

\paragraph{Gridworld MDP:} The gridworld depicted in Figure 6 is treated as a finite horizon undiscounted Markov decision process (MDP). States $s_t$ correspond to locations in the gridworld and actions $a_t$ to a choice of ‘going left’ or ‘going right’. Transitions are stochastic, such that each action results in the desired transition only some percentage of the time; otherwise, the agent transitions downward. The probabilities of an (unlucky) downward transition for the rightward and leftward action were 0.08 and 0.04, respectively. The probability of transitioning from state $s_t$ to $s_{t+1}$ after taking action $a_t$ is denoted by $p(s_{t+1}|s_t,a_t)$. Three states are associated with a non-zero reward or cost: ‘Goal’, ‘Quit’ and ‘Lavapit’ with rewards/costs of $[+3,-2,-15]$, respectively. The reward/cost function is denoted as $r(s_t)$. Visitation to any of these three states ends the episode; i.e., from them, the agent transitions deterministically to a terminal state, which has a reward of $0$, and where it remains indefinitely. Otherwise, the episode terminates at horizon $T$.

\paragraph{Precommitted CVaR (pCVaR)} The optimal policy for the precommitted approach to CVaR (or pCVaR) was calculated using a modified version of the dynamic programming algorithm in \citep{chow2015risk}; we adapt their algorithm to the finite horizon setting for an easier comparison with the fixed approach (fCVaR).

Starting at time point $t\!=\!T\!-\!1$ and working backwards, the agent calculates a set of Q-values $Q_t(s_t,a_t,\alpha_t)$ for all states $s_t$, actions  $a_t$ and risk preferences $\alpha_t$: 

\begin{equation}
\label{eq_pCVaR_Q_update}
 Q_t(s_t,a_t,\alpha_t) = r(s_t) + \min_{\xi(s_{t+1}) \in \mathcal{U}(\alpha_t)} \sum_{s_{t+1}}p(s_{t+1}| s_t,a_t)\xi(s_{t+1})  V_{t+1}(s_{t+1},\alpha_t \xi(s_{t+1}))
\end{equation}

The state values $ V_t(s_t,\alpha_t)$ are updated by taking a maximum over Q-values,

\begin{equation}
\label{eq_pCVaR_V_update}
V_t(s_t,\alpha_t) = \max_{a_t} Q_t(s_t,a_t,\alpha_t), 
\end{equation}

or are set equal to 0 at the horizon ($t\!=\!T$). 

These update equations differ in two notable ways from traditional dynamic programming, which uses a risk-neutral objective. First, the transition probabilities are weighted by $\xi(s_{t+1})$, which are chosen to minimize the next state's value as much as possible, under the constraints that the individual weights are less than $\sfrac{1}{\alpha_t}$ and that the distorted probabilities still sum to 1. These two conditions are denoted by:

\begin{equation}
\label{eq_U_conditions}
\mathcal{U}(\alpha_t) \coloneqq \left\{ \xi: \xi(s_{t+1}) \in \left[0,\sfrac{1}{\alpha_t}\right], \sum\nolimits_{s_{t+1}} p(s_{t+1}| s_t,a_t)\xi(s_{t+1})  = 1 \right\} 
\end{equation}

Second, the value at risk preference $\alpha_t$ at time point $t$ (on the left-hand side of equation \ref{eq_pCVaR_Q_update}), is a function of the next states' values at potentially different risk-preference levels, given by $\alpha_{t+1} \leftarrow \xi(s_{t+1}) \alpha_t$. The first modification comes from the dual representation of CVaR as a distorted (or $\xi$-weighted) expectation \citep{artzner1999coherent}\footnote{The dual representation is $\text{CVaR}_\alpha [Z] = \min\nolimits_{\xi \in \mathcal{U}(\alpha)} E_\xi[Z] $}, and the second modification from the conditional decomposition theorem of CVaR in \citep{pflug2016time}. The consequence of these modifications is that $V_t(s_t,\alpha_t)$ equals the CVaR$_{\alpha_t}$ of the future sum of rewards (i.e. the return) starting in state $s_t$ and following the optimal policy from thereon. Since is $\alpha_{t+1} $ is continuous, $V_t$ and $Q_t$ are linearly interpolated using 21 log spaced points for $\alpha_t$, following \citep{chow2015risk}, and thus the values are only approximately equal to the CVaRs. Chow, et al. (2015) interpolate $\alpha V$ and show that this allows them to bound interpolation error in a value-iteration approach; for our finite time horizon simulations, we simply interpolate $V_t$ since we are not iterating the equations until convergence. 

The optimal policy $\pi_t(a_t|s_t,\alpha_t)$ is calculated by taking the action associated with the maximum Q-value in each state (i.e. as the argmax in equation \ref{eq_pCVaR_V_update}). For each episode, agent starts at a preferred level of risk $\alpha_0$ (which is set to $0.18$ for the simulations in Figure 6) and chooses an action according to $\pi_0(a_0|s_0,\alpha_0)$. However, it then needs to adjust this risk preference depending on the state transitions (or rewards/costs) that occur, for instance, $\alpha_{t+1} \leftarrow \xi(s_{t+1}) \alpha_t$ for a particular transition involving $s_{t+1}$. These weights $\xi(s_{t+1})$ used for these adjustments are obtained from equation \ref{eq_pCVaR_Q_update}.

\paragraph{Nested CVaR (nCVaR)} The optimal policy for the nested approach was calculated based on \citep{ruszczynski2010risk}. For all states $s_t$ and actions $a_t$ (and for a fixed, preferred risk preference $\bar\alpha$), the Q-values are calculated as:

\begin{equation}
\label{eq_nCVaR_Q_update}
 Q_{\bar\alpha,t}(s_t,a_t) = r(s_t) + \min_{\xi(s_{t+1}) \in \mathcal{U}(\bar\alpha)} \sum_{s_{t+1}}p(s_{t+1} | s_t,a_t)\xi(s_{t+1})  V_{\bar\alpha,t+1}(s_{t+1}) 
\end{equation}

For each state, the values are calculated as:

\begin{equation}
\label{eq_nCVaR_V_update}
V_{\bar\alpha,t}(s_t) = \max_{a_t} Q_{\bar\alpha,t}(s_t,a_t)
\end{equation}

The optimal policy $\pi_{\bar\alpha,t}(a_t|s_t)$ is again calculated by taking the actions that are associated with the maximum Q-value in each state (i.e the argmax in equation \ref{eq_nCVaR_V_update}).

Note that unlike for pCVaR, the values and policy are only represented and updated at a single, fixed risk preference ($\bar\alpha$). We denote this using a subscript to emphasize the difference. Also note that the $Q_t$ and $V_t$ no longer correspond to the CVaR of the return, like in the pCVaR case, but rather the nested sum of immediate rewards and future CVaR evaluations, as given in equation 4 in the main text. 

\paragraph{Fixed CVaR (fCVaR)} Calculating the optimal policy for the fixed CVaR approach involves a combination of the precommitted and nested approaches. 

At its heart, fCVaR involves a form of distributional RL in which the distribution over Q-values is represented using a set of CVaRs across $\alpha$ levels (again, using 21 possible values) rather than using quantiles \cite{dabney2018distributional}. Thus, for all states $s_t$, actions $a_t$ and risk preferences $\alpha_t$,  Q-values across $\alpha$ levels are updated using the same equation \ref{eq_pCVaR_Q_update} as the precommitted approach. In terms of distributional reinforcement learning, this corresponds to a distributional back-up.

Diverging from the pCVaR approach, however, and more like nCVaR, the (distributional) value for each state $s_t$ (still represented across multiple $\alpha_t$ values), is set according to the action that maximizes the Q-value for that state at a fixed, preferred level of risk ($\bar\alpha$):

\begin{align}
  \label{eq_fCVaR_V_update}
V_t(s_t,\alpha_t) &= Q_t(s_t,a^*_{\bar\alpha,t},\alpha_t), 
  \intertext{where $a^*_{\bar\alpha,t}$ is chosen to maximize:}
\label{eq_fCVaR_a_update}
a^*_{\bar\alpha,t} &= \argmax_{a} Q_t(s_t,a,\bar\alpha)
\end{align}

This alternates between faithfully backing up the distribution and choosing an action at each time step that greedily maximizes the CVaR of the future return at a fixed, preferred alpha level. As we discuss in the main text, this greedy action selection can lead to time inconsistency. Thus similarly to nCVaR, the optimal policy $\pi_{\bar\alpha,t}(a_t|s_t)$ is only a function of state $s_t$, with the preferred risk preference $\bar\alpha$ kept constant across time (yet potentially differing across agents or simulations).

\paragraph{Depicting the optimal policies:} Given the finite-horizon nature of the problem, the optimal policy for each state could potentially differ across time. Furthermore for pCVaR, the optimal policy can also differ depending on the risk-preference $\alpha_t$, which is adjusted after the start of the episode depending on the transitions that occur. Therefore, we used simulations to explore how the optimal policy unfolded across time for each agent. The agents started in state $s_0$ (upper left hand corner) and chose actions until they reached either the ‘Quit’, ‘Goal’, or ‘Lavapit’ state (this always occurred before $t=10$). In all of the simulations (20,000 for each agent), the action taken in each state was the same, regardless of the time point. This was true even for the pCVaR agent, which sometimes visited different alpha levels within a state (depending on how it got there). The actions from these simulations were plotted in Figure 6. The optimal policies for other levels of $\alpha$ are further explored in Supplemental Figure \ref{fig_gridworld_other_alphas}.

\FloatBarrier

\begin{figure}[h]
  \centering
  \includegraphics[width=1.0\linewidth]{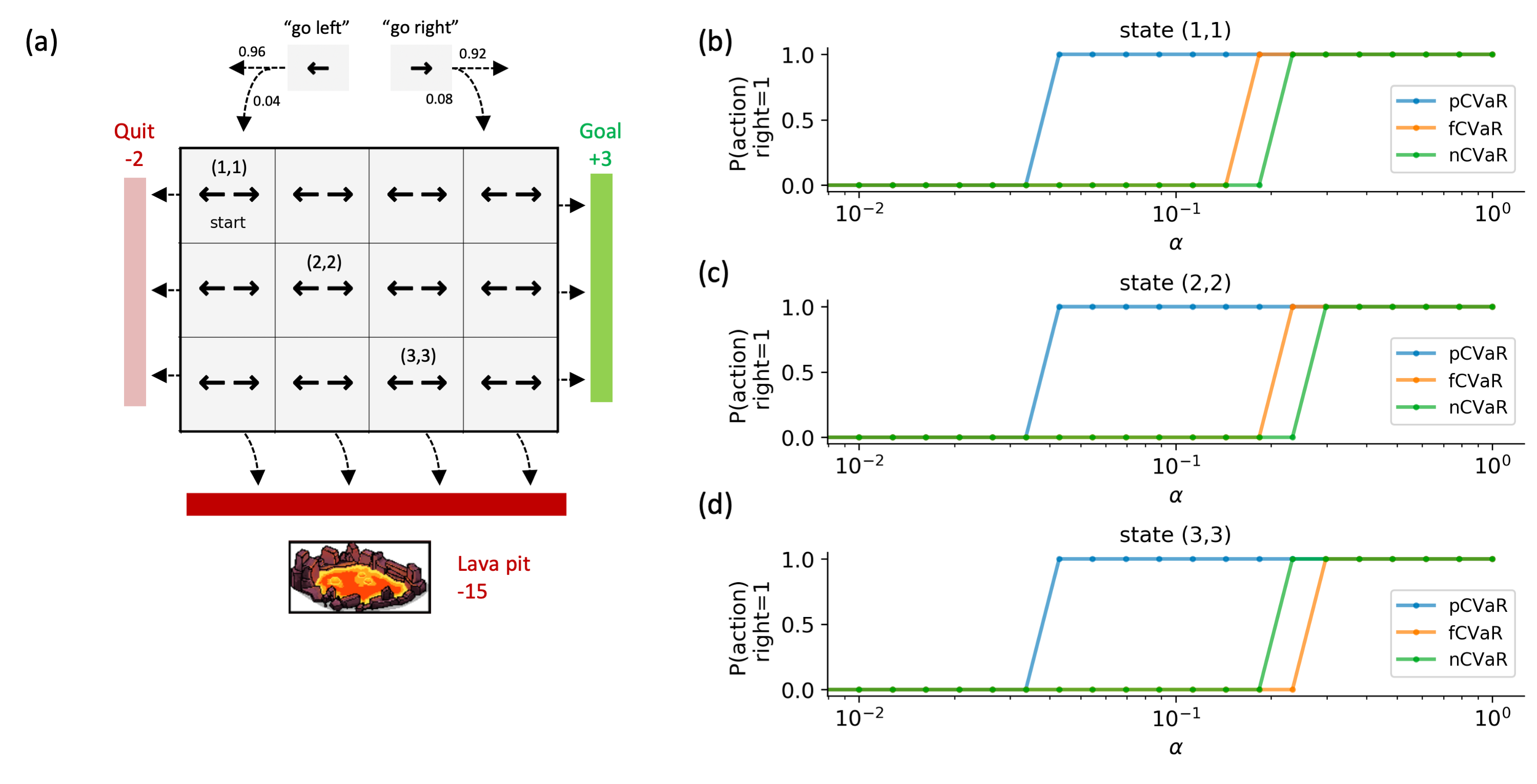}
  \caption{Optimal policies at other $\alpha$ levels. (a) Shows the gridworld MDP (also shown in Figure 6 in the main text). (b-d) The optimal policy is plotted as a function of $\alpha$ for pCVaR, fCVaR and nCVaR for three states: $(1,1)$, $(2,2)$, $(3,3)$, whose locations are labeled in panel a. The optimal policy was examined by running simulations starting in state $(1,1)$. The probability of taking a rightward action in simulation is plotted (this probability was either $0$ or $1$, as the agents always took the same action in each state). If a state was not visited during any of the simulations, the policy for the agent starting in that state is plotted (e.g. for states $(2,2)$ and $(3,3)$ at low alphas, where the agent decides to quit from the start). All three methods choose to go right towards the ‘Goal’ state at high levels of $\alpha$ and left towards the ‘Quit’ state at low levels of $\alpha$. The three methods differ, however, in when they choose to make this switch. pCVaR chooses to go right towards the goal at much lower levels of $\alpha$ than the other two methods in all three states. nCVaR requires the highest levels of $\alpha$ before it chooses right in states $(1,1)$ and $(2,2)$, reflecting its more conservative risk evaluation of distal threats (i.e. the lavapit). However, for state $(3,3)$, which is adjacent to the lavapit, it chooses to go right at a lower $\alpha$ level than fCVaR (but still a higher level than pCVaR). The 21 interpolation points used for $\alpha$ are shown as dots plotted on the lines.}
  \label{fig_gridworld_other_alphas}
\end{figure}

\bibliography{references}

\section*{Checklist}

\begin{enumerate}

\item For all authors...
\begin{enumerate}
  \item Do the main claims made in the abstract and introduction accurately reflect the paper's contributions and scope?
    \answerYes{}
  \item Did you describe the limitations of your work?
    \answerYes{}
  \item Did you discuss any potential negative societal impacts of your work?
    \answerNA{As our work investigates the psychology of decision making under risk, we do not anticipate it to have negative societal impacts.}
  \item Have you read the ethics review guidelines and ensured that your paper conforms to them?
    \answerYes{}
\end{enumerate}

\item If you are including theoretical results...
\begin{enumerate}
  \item Did you state the full set of assumptions of all theoretical results?
    \answerNA{}
	\item Did you include complete proofs of all theoretical results?
    \answerNA{}
\end{enumerate}

\item If you ran experiments...
\begin{enumerate}
  \item Did you include the code, data, and instructions needed to reproduce the main experimental results (either in the supplemental material or as a URL)?
    \answerNo{Data from the two-step task was already made publicly available at (https://osf.io/usdgt/) by the authors of the original study \citep{gillan2016characterizing}. The code used for the analyses/simulations will be made available upon request.} 
  \item Did you specify all the training details (e.g., data splits, hyperparameters, how they were chosen)?
    \answerYes{Critical details for the analyses and simulations were provided in the text. Additional information will be provided in the supplemental materials.}
	\item Did you report error bars (e.g., with respect to the random seed after running experiments multiple times)?
    \answerYes{}
	\item Did you include the total amount of compute and the type of resources used (e.g., type of GPUs, internal cluster, or cloud provider)?
    \answerNo{All analyses and simulations were run on a single, personal machine with limited compute.}
\end{enumerate}

\item If you are using existing assets (e.g., code, data, models) or curating/releasing new assets...
\begin{enumerate}
  \item If your work uses existing assets, did you cite the creators?
    \answerYes{}
  \item Did you mention the license of the assets?
    \answerNA{The data was made publicly available without license at https://osf.io/usdgt/}
  \item Did you include any new assets either in the supplemental material or as a URL?
    \answerNo{}
  \item Did you discuss whether and how consent was obtained from people whose data you're using/curating?
    \answerNA{Since we are reanalyzing data from \citep{gillan2016characterizing} we did not discuss this. Details about the consent process can be found in the original work.} 
  \item Did you discuss whether the data you are using/curating contains personally identifiable information or offensive content?
    \answerNA{} 
\end{enumerate}

\item If you used crowdsourcing or conducted research with human subjects...
\begin{enumerate}
  \item Did you include the full text of instructions given to participants and screenshots, if applicable?
    \answerNA{Details about the data collection process can be found in the original work \citep{gillan2016characterizing}.}
  \item Did you describe any potential participant risks, with links to Institutional Review Board (IRB) approvals, if applicable?
    \answerNA{See previous answer.} 
  \item Did you include the estimated hourly wage paid to participants and the total amount spent on participant compensation?
    \answerNA{See previous answer.} 
\end{enumerate}

\end{enumerate}
